# PGCN: Progressive Graph Convolutional Networks for Spatial–Temporal Traffic Forecasting

Yuyol Shin, *Member, IEEE*, and Yoonjin Yoon, *Member, IEEE*

***Abstract*— The complex spatial-temporal correlations in transportation networks make the traffic forecasting problem challenging. Since transportation system inherently possesses graph structures, many research efforts have been put with graph neural networks. Recently, constructing adaptive graphs to the data has shown promising results over the models relying on a single static graph structure. However, the graph adaptations are applied during the training phases and do not reflect the data used during the testing phases. Such shortcomings can be problematic especially in traffic forecasting since the traffic data often suffer from unexpected changes and irregularities in the time series. In this study, we propose a novel traffic forecasting framework called Progressive Graph Convolutional Network (PGCN). PGCN constructs a set of graphs by progressively adapting to online input data during the training and testing phases. Specifically, we implemented the model to construct progressive adjacency matrices by learning trend similarities among graph nodes. Then, the model is combined with the dilated causal convolution and gated activation unit to extract temporal features. With residual and skip connections, PGCN performs the traffic prediction. When applied to seven real-world traffic datasets of diverse geometric nature, the proposed model achieves state-of-the-art performance with consistency in all datasets. We conclude that the ability of PGCN to progressively adapt to input data enables the model to generalize in different study sites with robustness.**

***Index Terms*— Spatial-temporal traffic forecasting, graph neural networks, adaptive graph convolution, transportation networks, multivariate time-series.**

## I. Introduction

THE traffic forecasting problem has long been studied as a crucial technical capability of intelligent transportation systems (ITS), with the aim of predicting future traffic states of transportation networks using historical observations. The ability to accurately predict traffic patterns and conditions forms the basis for informed decision-making in traffic management, playing a vital role in alleviating congestion, enhancing safety, and optimizing traffic flow. The precise traffic prediction can have a multitude of real-world applications such as travel time estimation [1] and prospective traffic navigation [2]. The integration of accurate traffic forecasting into ITS is crucial for smart, efficient, and safe transportation networks. However, the complexity of spatial-temporal correlations in traffic data presents a challenge to achieving accurate forecasting.

Traditionally, statistical models, such as ARIMA (Auto-Regressive Integrated Moving Average) [3] and VAR (Vector Auto-Regressive) [4], and data-driven machine learning models, such as K-Nearest Neighbor [5] and Support Vector Regression [6], have been used for traffic forecasting. While these models have the advantage in interpretability of model parameters, they struggle to capture complex spatial-temporal correlations in traffic data.

In recent years, deep learning models, such as Recurrent Neural Networks (RNNs) and Convolutional Neural Networks (CNNs), have demonstrated the ability to increase the scope and accuracy of traffic forecasting. Recurrent Neural Networks (RNN) have naturally gained popularity for their ability to process sequential data [7], [8], [9], [10]. However, the gradient vanishing problem inherent in RNNs made them difficult to capture long-term relationships and the sequential computation of the models required prolonged training time. To overcome such limitations, convolutions have been introduced to extract temporal features [11], [12], [13], [14]. Excluding the sequential computation, convolutions were able to learn temporal correlations without the gradient vanishing problem with less training time. The recently developed self-attention mechanism is also used in traffic forecasting problems [15], [16], [17].

For spatial feature extraction, earlier works relied on CNNs, but they are limited by their inability to reflect the topological structure of transportation networks [11], [18], [19]. As a result, Graph Neural Networks (GNNs), which extract spatial features by aggregating information from adjacent nodes in transportation networks, have become a popular choice [8], [9], [10], [12], [13], [14], [15], [16], [17]. The state-of-the-art models for spatial-temporal traffic forecasting use GNNs to not only aggregate information from physically adjacent nodes but also construct adaptive graphs based on data-driven methods. STF-GNN [20] constructs an adaptive graph using Dynamic Time Warping (DTW) distances [21] using training data. Graph WaveNet [13] makes the model learn node embeddings during the training phase and constructs the self-adaptive graph. DMSTGCN [14] captures time-varying spatial correlations by constructing dynamic graphs for each time slot of a day. However, these models define the adaptive graphs using learnable node embeddings finalized during the training

Manuscript received 1 April 2023; revised 10 October 2023; accepted 16 December 2023. This work was supported by National Research Foundation of Korea (NRF) grants, funded by the South Korean Government (Ministry of Science and ICT), in part under Grant No. 2020R1A2C2010200 and in part under Grant No. 2021R1A4A1033486. The Associate Editor for this article was J. Sanchez-Medina. *(Corresponding author: Yoonjin Yoon.)*

The authors are with the Department of Civil and Environmental Engineering, Korea Advanced Institute of Science and Technology, Daejeon 34141, South Korea (e-mail: yuyol.shin@kaist.ac.kr; yoonjin@kaist.ac.kr).

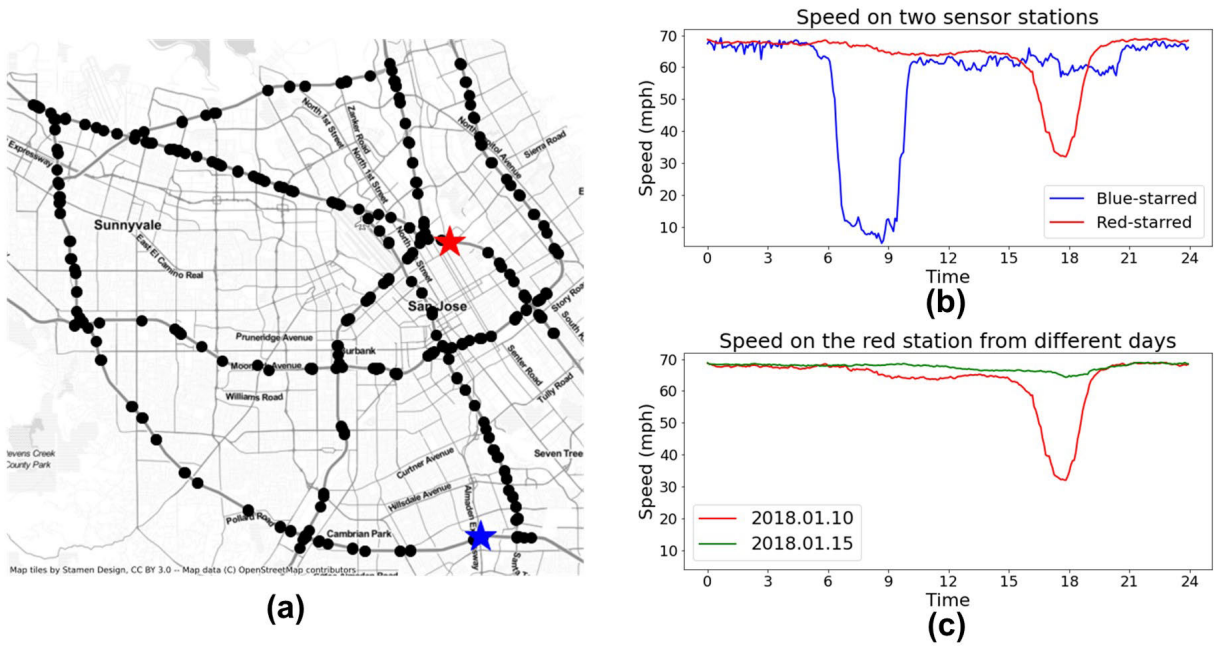

Fig. 1. Example of time-varying spatial correlation. (a) sensor stations near San Jose, CA, USA (*extracted from PeMS-Bay dataset explained in Section V-A*.); (b) Traffic speed from the selected (red star) sensor stations on the same day. Although not adjacent, they show similar speed trends during off-peak hours. (c) The speed trends on one sensor could change for the same time-of-day.

phase and do not adapt to the online traffic data for the testing phase.

The correlations between nodes can change over time for the same pair of nodes and for the same time-of-days. For example, two traffic nodes near a school and an office may have similar morning peak hours, leading to high spatial correlation between those nodes. On the other hand, the near-school node and the near-office node may experience different afternoon peak hours, which results in low spatial correlation for the entire time after the morning peak. In Fig. 1(b), real-world traffic speed data from two sensors are plotted. Although the two sensors are not adjacent to each other, they share similar trends during off-peak hours. However, the two nodes exhibit different peak hours (9 am and 6 pm), leading to weak correlations during those hours. Moreover, such observation does not always repeat for the same time-of-day. This may due to some other factors such as day-of-week, holidays, weather, accidents, or sports events. The plot in Fig. 1(c) shows that speed trends on a sensor could change for the same time-of-day. A dynamic graph based on online traffic data is necessary to capture the time-varying spatial correlation and to aggregate information from appropriate nodes.

In this study, we present a novel spatial-temporal traffic forecasting model, Progressive Graph Convolutional Networks (PGCN). We first define the trend similarities between traffic node signals using parameterized cosine similarity. The graph convolution module in the model constructs a progressive graph based on trend similarities. Although the parameters for the model are learned during the training phase, the model can adapt to the online input graph signals at both the training and testing phases by constructing dynamic graphs based on input signals. Then, a graph convolution is operated on progressive graphs and transition matrices. In this way, the graph convolution module can capture spatial correlation among the nodes considering both the geographical adjacency and the online spatial correlation. Also, we implement dilated causal convolution [22] to extract temporal features of traffic data. The main contributions of our paper are:

- We propose a novel spatial-temporal traffic forecasting model, Progressive Graph Convolution Network (PGCN) of which the main components are the progressive graph convolution module and the dilated causal convolution. The progressive graph convolution module extracts spatial features by capturing time-varying spatial correlation, and dilated causal convolution is implemented with the gated activation unit to extract temporal features.
- Progressive graphs are constructed for the progressive graph convolution module. These graphs capture the time-varying spatial correlation using the parameterized cosine similarity between node signals. The graphs update the edges and weights at each time step by adapting to online input data.
- Experiments on seven real-world datasets demonstrate that the proposed model consistently achieves state-of-the-art performance. While the other baseline models fail to perform on specific datasets, PGCN keeps its performance within the 1.5% range of the best-performing model except for few cases. In addition, the suggested model shows competitiveness for predictions without information on the structural network.[1]

The rest of this article is organized as follows. In Section II, we introduce the studies related to this article, focusing on the development of spatial-temporal feature extraction methods for traffic forecasting. Sections III and IV explain the preliminaries for our study and the proposed model. Section V describes the datasets, experimental setting, and methods for comparison along with the results of experiments. In Section VI, we present the conclusion of the study and the directions of our future work.

## II. Related Works

### A. Graph Neural Networks

Graph Neural Networks have been applied in a variety of research areas including protein structure prediction, citation network [23], recommender system [24], and action recognition [25] to learn graph-structured data. Also, in spatial-temporal forecasting, the GNNs have been a popular choice to extract spatial features. The networks under the umbrella of graph neural networks can be categorized into three groups - the convolutional, attentional, and message-passing GNNs [26] - having differences in how they aggregate information from neighboring nodes. The convolutional GNNs [8], [27], [28] multiply a constant value to the source node features and conduct aggregating operations. The group of spectral graph convolutions constructs filter in the Fourier domain and extracts spatial features [27], [28]. There also is a group of spatial-based graph convolutions that defines the convolution based on the spatial relations among the nodes. GraphSAGE [29] samples a fixed number of neighbor nodes for each node and aggregates the information. Diffusion Convolution [8] defines the probability transition matrix for the nodes and regards the information propagation in the graphs as a diffusion process. The attentional GNNs calculate the multiplier with a function of the source node and target node features. Graph Attention Networks [23] implements the multi-head attention mechanism by using the concatenation of source and target node features as queries, learnable parameters as keys, and source node features as values. GaAN [30] implements the attention method with an additional gate to control the information of neighbor nodes. The last category, message-passing computes the vector-based features as a function of the source and target node features [31], [32].

[1] Code and dataset are available at https://github.com/yuyolshin/PGCN.

### *B. GNNs for Spatial-Temporal Traffic Forecasting*

While the most common form of graph neural networks takes the adjacency matrices with only structural connectivity information, there have been several efforts to incorporate more structural information in traffic forecasting. Li et al. [8] suggested to crop edges based on the geospatial distance among the nodes, reflecting the distance information of the transportation networks. DDP-GCN [33] showed incorporating various structural information such as distance, heading direction and joint angle could enhance the forecasting power of the models, and the MW-TGC network [10] showed that different structural characteristics should be considered in different scenarios. However, these methods have all relied on static information such as the distances and joint angles between node pairs and the speed limits of road segments, while nodes might share similar characteristics without being physically close or connected in transportation networks. To reflect relationships beyond physical connection, there was an effort to construct semantic graphs based on functional similarity, and transportation connectivity [34]. Li and Zhu [20] proposed an approach to construct graphs based on input data using Dynamic Time Warping distance [20]. Several works [13], [14], [35], [36] showed learning an adaptive graph during the training phase could enhance the performance of the model even further. Graph WaveNet [13] constructed an adaptive adjacency matrix by multiplying self-learned node embeddings, and DMSTGCN [14] showed constructing one adaptive graph for each time slot of a day could enhance the performance even further. Bai et al. [36] suggested a model that generated node-specific parameters as well as adaptive graphs based on data. However, these methods have defined the graphs before the validation and testing phases. Several methods have elaborated on updating connectivity information during testing phases [37], [38]. Z-GCNETs [37] used zigzag persistence images to capture the underlying topology of dynamic traffic data that persisted over time, and ASTGNN [38] extracted spatial correlation matrices utilizing attention outputs to update the weights of the graphs during testing time. Since the trend of spatial-temporal data may confront changes in daily trends and other unexpected situations during testing time, there needs a method to adapt to the online input data in both the training and testing phases. In this study, we suggest the progressive graph convolution that can update the edge connections of traffic graphs as well as the weights of the edges.

### *C. Temporal Feature Extraction*

While GNNs have become a popular choice of spatial feature extraction, other approaches have been implemented for temporal feature extraction for spatial-temporal traffic forecasting. In the initial deep models for traffic forecasting, RNNs have been a popular choice as they naturally possess the ability to learn sequential data [7], [8], [9], [10]. Another line of work has implemented temporal convolution [11], [12], which has been much lighter in terms of computation expense. However, these implementations have required deeper model structures to increase the size of the reception field and dilated convolutions have been suggested to overcome such limitations [13], [14]. As self-attention has made a large advancement in the field of natural language processing recently, a few studies have implemented the attention mechanism [15], [16]. There have been studies where temporal features have been extracted by constructing graphs incorporating temporal connectivity and conducting graph convolution on the graphs [20], [39].

## III. PRELIMINARIES

### *A. Notations and Definitions*

*Definition 1 (Transportation Network Graph):* We represent the transportation network graph as a directed graph $\mathcal{G} = (V, E)$, where $V$ is a set of $N$ nodes and $E$ is a set of edges representing pairwise connections between the nodes. An adjacency matrix $\boldsymbol{A} = (A_{ij}) \in \mathbb{R}^{N \times N}$ is a square matrix where the element $A_{ij}$ represent the weight of an edge $(v_i, v_j) \in E$.

*Definition 2 (Graph Signal):* The signal from node $v_i$ at time $t$ is denoted as $\boldsymbol{x}_t^i \in \mathbb{R}^C$ where $C$ is the number of input features. The graph signal is denoted as $\boldsymbol{X}_t = [\boldsymbol{x}_t^1, \boldsymbol{x}_t^2, \ldots, \boldsymbol{x}_t^N] \in \mathbb{R}^{N \times C}$. The signals of node $v_i$ observed during the last $T$ time steps from time $t$ are denoted $\boldsymbol{x}_t^{i^{(T)}} \in \mathbb{R}^{T \times C}$. Similarly, $\boldsymbol{X}^{i^{(T)}} \in \mathbb{R}^{T \times N \times C}$ is the graph signal for the entire graph.

### *B. Problem Definition*

The traffic forecasting problem is to predict future traffic states using historical traffic states such as speed, flow, and occupancy. Given the historical data of $T$ time steps from time $t$, the problem is to find function $H$ such that:

$$H : [\boldsymbol{X}_{t-T+1}, \boldsymbol{X}_{t-T+2}, \ldots, \boldsymbol{X}_t; \mathcal{G}] \rightarrow [\hat{\boldsymbol{Y}}_{t+1}, \hat{\boldsymbol{Y}}_{t+2}, \ldots, \hat{\boldsymbol{Y}}_{t+T'}] \tag{1}$$

where $\hat{\boldsymbol{Y}}_{t+k}$ is the predicted traffic states at time $t+k$, $(1 \leq k \leq T')$, and $T'$ is the number of time steps to predict.

## IV. PROGRESSIVE GRAPH CONVOLUTIONAL NETWORKS

In this section, we explain the core idea of Progressive Graph Convolutional Networks (PGCN). We first explain the procedure for constructing a progressive graph, and graph convolution operation using a progressive adjacency matrix. Then, we describe the implementation of dilated causal convolution combined with the graph convolution module. Lastly, the overall framework of PGCN is illustrated.

### *A. Progressive Graph Construction*

The similarity between two node signals changes over time. For example, while the morning peak hours near schools and office buildings are similar, the afternoon peak hours can be quite different. Modeling such correlations based on static features such as POI category or speed limit is intuitive, but it may not be the most efficient approach considering the scale and complexity involved. Instead, we propose to learn rich semantics hidden in the online traffic data itself by measuring node similarities beyond the simple spatial adjacency.

In this study, we first define trend similarities between node signals using parameterized cosine similarity and propose the progressive graph that can progressively adapt to traffic change based on trend similarities. Wu et al. [13] and Han et al. [14] employed similar ideas to adaptively learn the adjacency matrices. However, their methods only updated the graph during the training phase and did not incorporate the online graph signals during the testing phase.

*Definition 3 (Progressive Graph, p-graph):* As the correlations between different nodes evolve, it is intuitive to update the node relationship progressively using online traffic data. Progressive graph ($p$-graph) $\{\mathcal{G}^t\}$ is a set of graphs where $\mathcal{G}^t = (V, \boldsymbol{A}^t_P)$. Here, $\boldsymbol{A}^t_P$ is the progressive adjacency matrix at time $t$, which contains the pairwise weights learned from the node signal similarities.

The goal is to impose a higher weight between nodes with similar signals regardless of their spatial proximity. The trend similarities among nodes of a graph are measured using the cosine similarity of their signals. Suppose the node signal $\boldsymbol{x}^i_t$ has one input feature. The cosine similarity $s_{ij}$ between two nodes $v_i$ and $v_j$ is defined as:

$$s^t_{ij} = \tilde{\boldsymbol{x}}^{i^{(T)\top}}_t \cdot \tilde{\boldsymbol{x}}^{j^{(T)}}_t, \tag{2}$$

where $\tilde{\boldsymbol{x}}^{i^{(T)}}_t = \bar{\boldsymbol{x}}^{i^{(T)}}_t / \|\bar{\boldsymbol{x}}^{i^{(T)}}_t\|$ is a unit vector, and $\bar{\boldsymbol{x}}^{i^{(T)}}_t$ is the min-max normalized signal of node $v_i$ at time $t$. We normalized the node signals to consider the similarities in terms of trends rather than absolute values since two nodes can possess similar trends with different values. For example, if two nodes generate signal vectors as [20, 30, 20, 40, 20] and [50, 60, 50, 70, 50], the two signals have an identical range and trend and yield cosine similarity 1 for normalized signals, even if the magnitudes are different. In addition, it is a common practice in time-series analyses to rescale and normalize data for comparisons.

Furthermore, we implemented a learnable parameter $\boldsymbol{W}_{adj} \in \mathbb{R}^{T \times T}$ to the cosine similarity to adapt to the patterns and randomness of noises in real-world datasets. With $\boldsymbol{W}_{adj}$, each element of the progressive adjacency matrix $A^t_{P_{ij}}$ is defined

$$A^t_{P_{ij}} = \text{softmax}\left(\text{ReLU}\left(\tilde{\boldsymbol{x}}^{i^{(T)\top}}_t \boldsymbol{W}_{adj} \tilde{\boldsymbol{x}}^{j^{(T)}}_t\right)\right). \tag{3}$$

The softmax function is applied to normalize the progressive adjacency matrix, and ReLU activation eliminates the negative connections. The parameter $\boldsymbol{W}_{adj}$ learns the relationship between the two signals $\tilde{\boldsymbol{x}}^{i^{(T)}}_t$ and $\tilde{\boldsymbol{x}}^{j^{(T)}}_t$ after transforming each vector with $\boldsymbol{W}_{adj}$. In other words, $\boldsymbol{W}_{adj}$ encodes the linear transformations applied to the signals to obtain the final similarity value. For signals with more than one feature, similarities are calculated for each feature. If the signal has three features, three different progressive adjacency matrices can be defined. Note that the parameter $\boldsymbol{W}_{adj}$ is learned during the training phase, and online node signals $\tilde{\boldsymbol{x}}^{i^{(T)}}_t$ and $\tilde{\boldsymbol{x}}^{j^{(T)}}_t$ are used to progressively update the graphs.

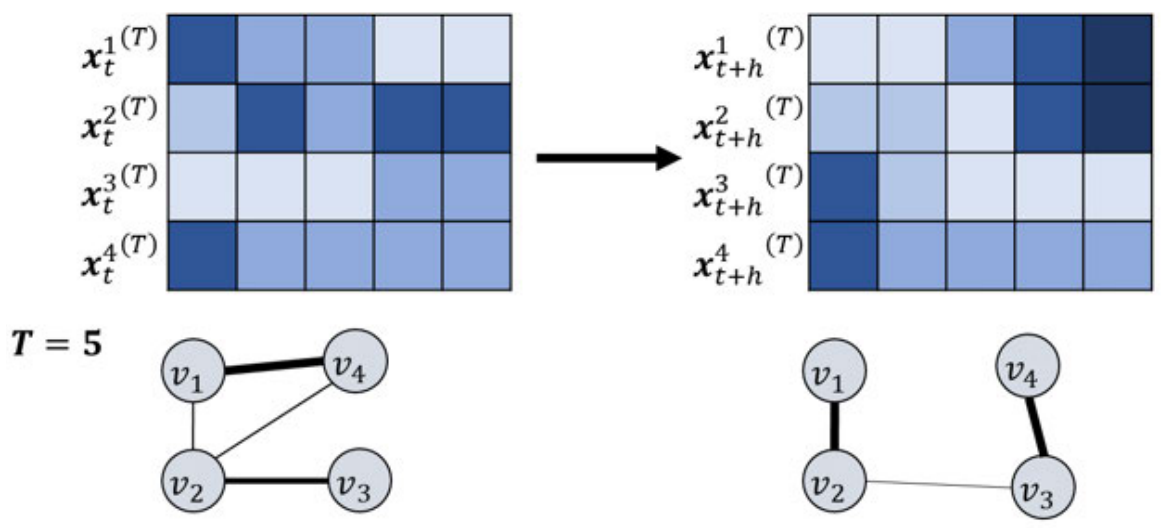

Fig. 2. Progressive Graph ($p$-graph). Each row in the matrix is the illustration of the node signal profile given the last 5 time steps. At $t$, $v_1$ shows the highest similarity with $v_4$, but such connection disappears at $t + h$.

Fig. 2 illustrates the idea of the progressive graph. Given four nodes $\{v_1, v_2, v_3, v_4\}$, each row in the matrix illustrates a single feature node signal observed in the last $T = 5$ time steps. We present two matrices for time $t$ and $t + h$. At $t$, one can observe the strong similarity between $v_1$ and $v_4$, whereas $v_1$ is most similar to $v_2$ at time $t + h$. It is also notable that edges can appear/disappear over time.

### B. Progressive Graph Convolution Module

The core idea of any graph convolution module is to aggregate neighbor nodes' information in extracting spatial features for the target node. The basic form of a graph convolution module is multiplying graph signal and learnable parameters to adjacency matrices processed by a defined method. In traffic forecasting, one of the most popular forms of graph convolution modules is Diffusion Convolution [8], in which traffic flow on a transportation network is considered a diffusion process. With the transition matrix $\boldsymbol{P} = \boldsymbol{A}/rowsum(\boldsymbol{A})$, the diffusion convolution on a directed graph for $K$-step diffusion process with filter $f_{\boldsymbol{W}}$ can be defined as

$$\boldsymbol{Z}_t = \boldsymbol{X}_t \star_{\mathcal{G}} f_{\boldsymbol{W}} = \sum_{k=0}^{K-1} \boldsymbol{P}^k \boldsymbol{X}_t \boldsymbol{W}_{k,1} + \boldsymbol{P}^{\top^k} \boldsymbol{X}_t \boldsymbol{W}_{k,2}, \tag{4}$$

where $\star_{\mathcal{G}} f_{\boldsymbol{W}}$ is the graph convolution operation with filter $f_{\boldsymbol{W}}$, and $\boldsymbol{W}_{k,1}$, and $\boldsymbol{W}_{k,2} \in \mathbb{R}^{C \times D}$ are learnable parameters. $\boldsymbol{P}$ and $\boldsymbol{P}^\top$ are used to reflect the forward and backward diffusion process. If the given adjacency matrix is undirected, only the first term of equation (4) is used.

Taking the diffusion convolution as the base graph convolution module for our progressive graph convolution, we added the multiplication of the progressive adjacency matrix, a graph signal matrix, and an additional weight parameter to diffusion convolution.

$$\boldsymbol{Z}_t = \boldsymbol{X}_t \star_{\mathcal{G}} f_{\boldsymbol{W}} = \sum_{k=0}^{K-1} \boldsymbol{P}^k \boldsymbol{X}_t \boldsymbol{W}_{k,1} + \boldsymbol{P}^{\top^k} \boldsymbol{X}_t \boldsymbol{W}_{k,2} + \boldsymbol{A}^t_P \boldsymbol{X}_t \boldsymbol{W}_{k,3}. \tag{5}$$

### C. Dilated Causal Convolution

In PGCN, we implement dilated causal convolution [40] to extract temporal features of graph signals. Causal convolutions extract temporal features by stacking 1-D convolution layers while ensuring that future information is not considered for prediction. This operation can be conducted as in Fig. 3(a), shifting the output features of each convolution layer by a few time steps. Since the main component of the causal convolution is convolution layers, it does not require sequential computation as in recurrent units. Compared to the self-attention layer, which has become another popular choice in time-series modeling, causal convolutions require a smaller number of learnable parameters, thus making a model more concise. Dilated causal convolution is suggested to overcome the limitation of causal convolution, requiring many number of layers to increase the size of the reception field. By skipping input values by defined dilation factor, the size of the reception field can increase exponentially with the number of hidden layers (Fig. 3(b)). Given a single feature time-series input $\boldsymbol{x}_t^{i(T)} \in \mathbb{R}^T$ at time $t$, and a convolution kernel $\gamma \in \mathbb{R}^P$, the dilated causal convolution applied on $\boldsymbol{x}_t^{i(T)}$ can be represented as

$$\boldsymbol{x}_t^{i(T)} \star_{\mathcal{T}} \gamma = \sum_{p=0}^{P} \gamma(p)\boldsymbol{x}_t^{i(T)}(t - d \times p), \tag{6}$$

where $\star_{\mathcal{T}}\gamma$ is a dilated convolution operation with kernel $\gamma$, $d$ is the dilation factor, and scalar values in the parentheses indicate the indices of the vectors. Finally, the temporal feature of the input sequence $\boldsymbol{X}_t^{(T)} \in \mathbb{R}^{N \times T \times C}$ is extracted by passing the input to gated activation units following the dilated causal convolution

$$\boldsymbol{H}_t = \tanh\left(\boldsymbol{X}_t^{(T)} \star_{\mathcal{T}} \Gamma_1\right) \odot \sigma\left(\boldsymbol{X}_t^{(T)} \star_{\mathcal{T}} \Gamma_2\right) \tag{7}$$

where $\Gamma_1$ and $\Gamma_2 \in \mathbb{R}^{P \times C \times D}$ are the kernels for dilated causal convolutions, $\odot$ denotes the element-wise multiplication, and $\sigma(\cdot)$ is a sigmoid activation function.

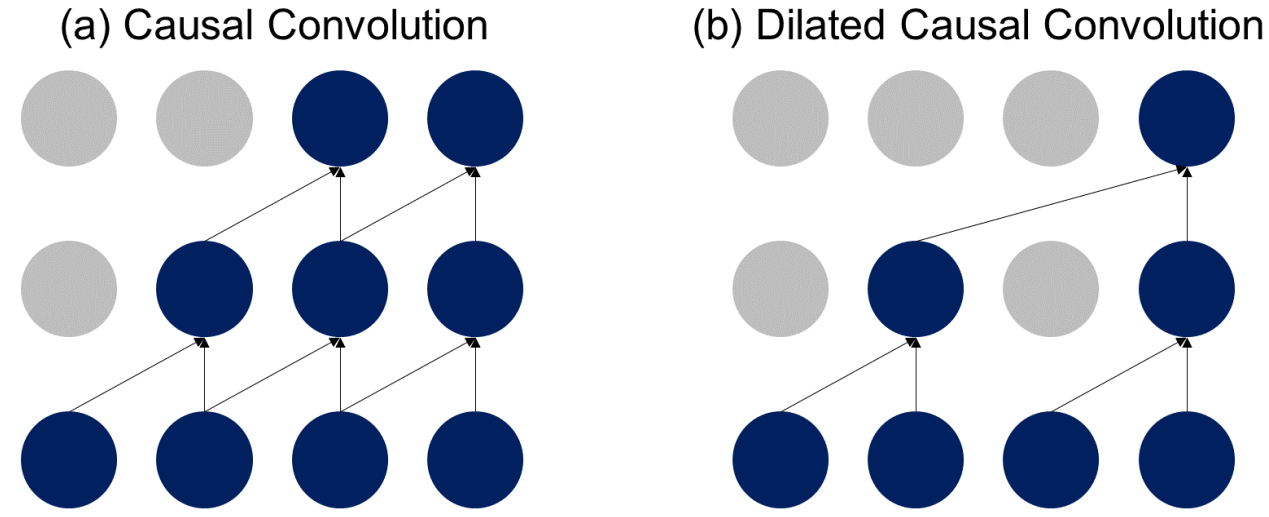

Fig. 3. Causal convolution and dilated causal convolution.

### *D. Overall Architecture of PGCN*

Fig. 4 shows the overall architecture of PGCN. Each spatial-temporal layer consists of dilated casual convolutions, a gated activation unit, a progressive graph convolution module, and a residual connection. While multiple spatial-temporal layers are staked to increase the size of the reception field in temporal feature extraction, the parameter for the progressive graph constructor is shared across the layers. Finally, the output layer consists of a skip connection from each layer to prevent information loss from the initial layers and two fully connected

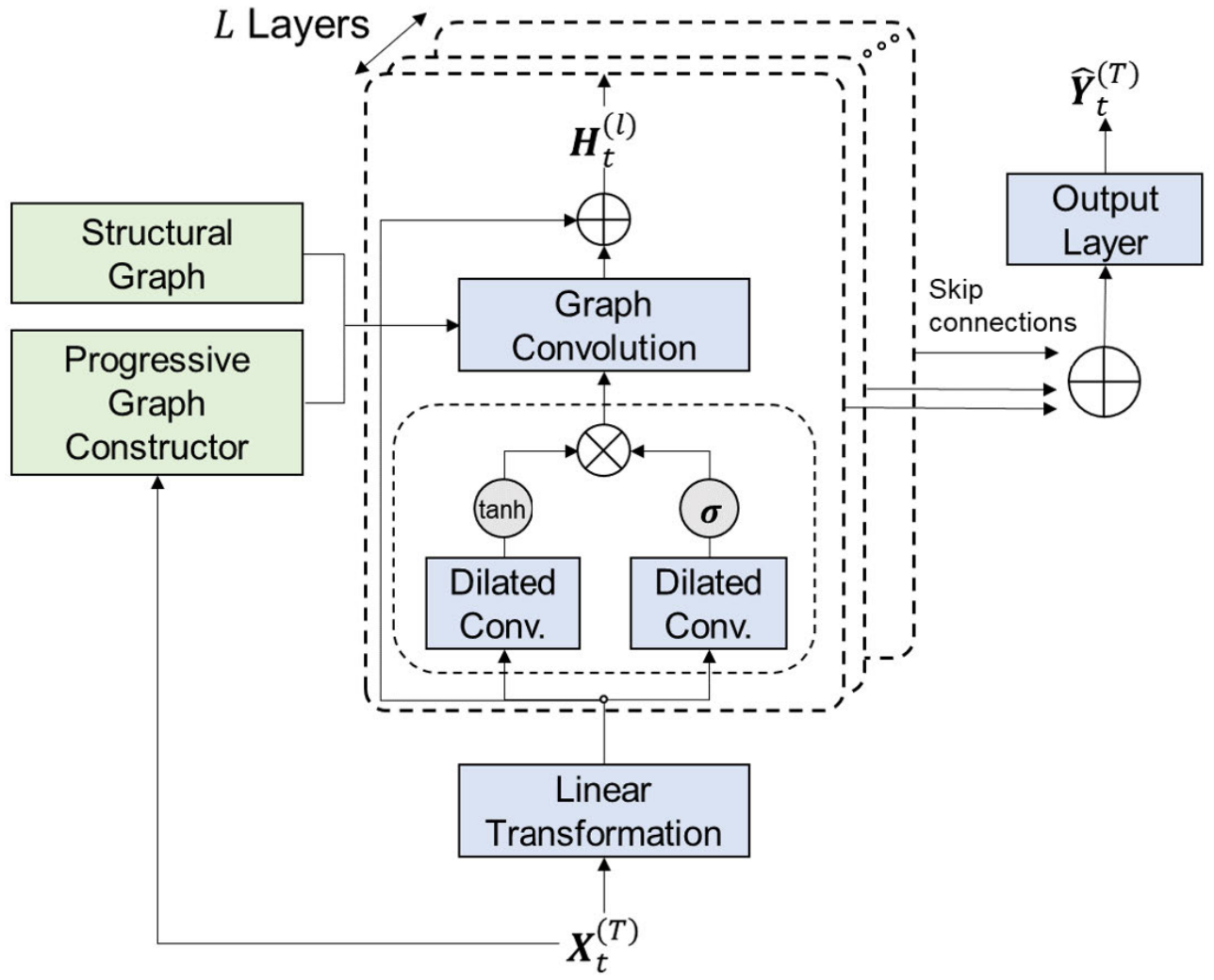

Fig. 4. The overall architecture of PGCN.

layers with ReLU activation. We used mean absolute error (MAE) for training.

$$\text{MAE}\left(\hat{\boldsymbol{Y}}_t^{(T)}, \boldsymbol{Y}_t^{(T)}\right) = \frac{1}{TN}\sum_{i=1}^{N}\sum_{j=0}^{T-1}\left|\hat{y}_{t-j}^{i} - y_{t-j}^{i}\right| \tag{8}$$

## V. Experiments

### *A. Datasets*

To evaluate the performance of the proposed model, we applied the model to seven real-world datasets, namely, PeMS-Bay, METR-LA [8], Urban-core [41], Seattle-Loop [9], PEMS04, PEMS08 [42], and PEMS07 [39].

PeMS-Bay is a highway speed dataset for traffic forecasting collected by California Transportation Agencies (CalTrans) Performance Measurement System (PeMS). The dataset includes six months of data aggregated in the frequency of 5 minutes. Spatially, it consists of 325 loop detectors in the Bay area are included.

METR-LA is a highway traffic flow dataset containing the data collected from 207 loop detectors in Los Angeles County. The dataset contains three months of data aggregated in the frequency of 5 minutes. For PeMS-Bay and METR-LA, we used the same data preprocessing procedures as in [8].

Urban-core is an urban speed dataset containing data collected from DTG (Digital Tacho Graph) on Seoul Taxis. The dataset includes speed data aggregated at the road segment level in a 5-minute resolution for one month. The number of road segments in the dataset is 304.

Seattle-Loop is a highway speed dataset collected from 323 loop detectors in the Greater Seattle Area. The dataset contains 5-minute resolution data for the entirety of 2015.

PEMS04 and PEMS08 contain the information collected from PeMS. Different from PeMS-Bay, these datasets include three features of traffic measures which are speed, traffic flow, and occupancy. The spatial-temporal coverage of PEMS04 is San Francisco Bay Area from 1 January 2018 to 28 February 2018. For PEMS08, it is the San Bernardino area from 1 July

TABLE I
SUMMARY OF THE DATASETS

| Datasets | # Samples | # Nodes | Frequency | Traffic |
|---|---|---|---|---|
| PeMS-Bay | 52,116 | 325 | 5 min | S |
| METR-LA | 34,272 | 207 | 5 min | F |
| Urban-core | 8,640 | 304 | 5 min | S |
| Seattle-Loop | 105,120 | 323 | 5 min | S |
| PEMS04 | 16,992 | 307 | 5 min | S, F & O |
| PEMS08 | 17,856 | 170 | 5 min | S, F & O |
| PEMS07 | 28,224 | 883 | 5 min | F |

S: Speed F: Traffic flow O: Occupancy

2016 to 31 August 2016. PEMS04 and PEMS08 are collected from 307, and 170, respectively.

We also use PEMS07, which is a highway flow dataset collected from 883 loop detectors in Los Angeles and Ventura counties. The dataset contains four months of data from 1 May 2017 to 31 August 2017 in a 5-minute resolution.

The training, validation, and test sets are divided in a proportion of 0.7, 0.1, and 0.2 for all datasets except Urban-core. For Urban-core, we used 21 days, 2 days, and 7 days division. For all datasets, we concatenated time-of-day information as an additional field. Table I contains the summary of the datasets.

### B. Experiment Settings

For metrics, we chose Mean Absolute Error (MAE), Root Mean Squared Error (RMSE), and Mean Absolute Percentage Error (MAPE). MAE is as defined in equation (8), and RMSE and MAPE can be defined as:

$$\mathrm{RMSE}\left(\hat{\boldsymbol{Y}}_t^{(T)}, \boldsymbol{Y}_t^{(T)}\right) = \sqrt{\frac{1}{TN}\sum_{i=1}^{N}\sum_{j=0}^{T-1}\left(\hat{y}_{t-j}^{i} - y_{t-j}^{i}\right)^2} \tag{9}$$

$$\mathrm{MAPE}\left(\hat{\boldsymbol{Y}}_t^{(T)}, \boldsymbol{Y}_t^{(T)}\right) = \frac{1}{TN}\sum_{i=1}^{N}\sum_{j=0}^{T-1}\frac{\left|\hat{y}_{t-j}^{i} - y_{t-j}^{i}\right|}{y_{t-j}^{i}} \tag{10}$$

For the experiment of PGCN, we implemented the model with 8 spatial-temporal layers and set the number of hidden dimensions to 32. For dilated causal convolutions, dilation factors are 1, 2, 1, 2, 1, 2, 1, and 2 with a convolution kernel size of 2. We used Adam optimizer for the training with an initial learning rate of 0.001. We ran the model for 100 epochs and chose the model that best performed in the validation set for evaluation. The model was trained in a Pytorch environment with one NVIDIA TITAN RTX with 24GB memory (GPU) and Intel(R) Xeon(R) CPU ES-2630 v4 @ 2.20GHz (CPU).

### C. Methods for Comparison

To evaluate the performance of the proposed model, we compared the experimental results with the following baseline models:

- HA (Historical Average) takes the traffic state of the most recent time step which is used to predict the following time steps.
- FC-LSTM combines LSTM units with a fully connected layer.
- DCRNN [8] combines the diffusion convolution layer with recurrent units to make the spatial-temporal prediction.
- Graph WaveNet [13] implements dilated causal convolution to extract temporal features and graph convolution with a self-adaptive adjacency matrix to extract spatial features.
- TGC-GRU uses Traffic Graph Convolution [9] with recurrent units (Free-flow matrices and additional loss terms are omitted in experiments in this study).
- DMSTGCN [14] constructs an adaptive graph for each time slot of a day and conducts graph convolution operation. For single feature datasets, we did not implement the layer for auxiliary information modeling.
- STSGCN [39] builds a large graph that captures the localized spatial-temporal correlations to synchronously extract both spatial and temporal features of traffic data, and avoid separate extraction of each feature.
- AGCRN [36] generates both node-specific parameters and adaptive graphs. The model combines the spatial feature extraction module with GRU to acquire spatial-temporal features.
- Z-GCNETs [37] adopts zigzag persistent diagrams (ZPD) to represent spatial-temporal traffic data. Zigzag persistent images and an adaptive graph convolution module is combined with GRU. We did not evaluate this model on Seattle-Loop due to the extensive computation time of calculating zigzag persistence images.

### D. Experimental Results

*1) Overall Performance Evaluation:* Table II shows the performance of the proposed model and the baselines on the seven datasets for 15, 30, and 60 minutes predictions. On all datasets, deep neural network (DNN) models outperform the historical average except for TGC-GRU on METR-LA, and TGC-GRU and FC-LSTM on Seattle-Loop and PEMS07 on 15-min prediction. DCRNN shows competitive performance on Seattle-Loop, and yields lower performances on PEMS04, PEMS08, and PEMS07. On Urban-core, FC-LSTM and TGC-GRU tend to show lower performance compared to the other DNN models for 15-min prediction, while for 60-min, Graph WaveNet fails to show competitive performance. On multi-feature datasets, PEMS04 and PEMS08, DMSTGCN becomes more competitive than its implementation on single-feature datasets. Taking an additional feature as auxiliary information, the model achieves the best performances for 15-min prediction (MAPE) on PEMS08, and 30-min (MAE, RMSE, MAPE) and 60-min (MAE, RMSE) predictions on PEMS04. AGCRN produces the best results for several metrics and prediction horizons on Urban-core, Seattle-Loop, and PEMS07.

PGCN achieves the best performance in at least one performance metric in all tasks except predictions for 15-min on METR-LA, 15-min on Seattle-Loop, 30-min on PEMS04, and 15-min on PEMS08. However, the model achieves the runner-up performance for all metrics in all these tasks except for 15-min MAPE on PEMS08. Although it does not always significantly outperform the other models in all datasets, our model is shown to be the model with the most consistent

TABLE II
FORECASTING OUTCOMES OF PGCN AND BASELINE MODELS ON SEVEN REAL-WORLD DATASETS

| | Model | 15 min | | | 30 min | | | 60 min | | |
|---|---|---|---|---|---|---|---|---|---|---|
| | | MAE | RMSE | MAPE | MAE | RMSE | MAPE | MAE | RMSE | MAPE |
| PeMS-Bay | HA | 1.60 | 3.43 | 3.24 | 2.18 | 4.99 | 4.65 | 3.05 | 7.01 | 6.83 |
| | FC-LSTM | 2.05 | 4.19 | 4.80 | 2.20 | 4.55 | 5.20 | 2.37 | 4.96 | 5.70 |
| | DCRNN | 1.38 | 2.95 | 2.90 | 1.74 | 3.97 | 3.90 | 2.07 | 4.74 | 4.90 |
| | Graph WaveNet | **1.30** | 2.74 | 2.73 | 1.63 | 3.70 | 3.67 | 1.95 | 4.52 | 4.63 |
| | TGC-GRU | 2.34 | 4.30 | 5.08 | 2.44 | 4.64 | 5.40 | 2.57 | 5.01 | 5.78 |
| | DMSTGCN | 1.33 | 2.83 | 2.80 | 1.67 | 3.79 | 3.81 | 1.99 | 4.54 | 4.78 |
| | STSGCN | 1.42 | 2.95 | 2.99 | 1.75 | 3.93 | 3.93 | 2.06 | 4.72 | 4.81 |
| | AGCRN | 1.39 | 2.96 | 2.98 | 1.71 | 3.92 | 3.87 | 2.02 | 4.69 | 4.70 |
| | Z-GCNETs | 1.36 | 2.86 | 2.88 | 1.68 | 3.78 | 3.79 | 1.98 | 4.53 | 4.60 |
| | PGCN | **1.30** | **2.73** | **2.72** | **1.62** | **3.67** | **3.63** | **1.92** | **4.45** | **4.45** |
| METR-LA | HA | 4.02 | 8.69 | 9.39 | 5.09 | 11.13 | 12.21 | 6.80 | 14.21 | 16.71 |
| | FC-LSTM | 3.44 | 6.30 | 9.60 | 3.77 | 7.23 | 10.90 | 4.37 | 8.69 | 13.20 |
| | DCRNN | 2.77 | 5.38 | 7.30 | 3.15 | 6.45 | 8.80 | 3.60 | 7.60 | 10.50 |
| | Graph WaveNet | **2.69** | **5.15** | **6.90** | **3.07** | **6.22** | **8.37** | **3.53** | 7.37 | 10.01 |
| | TGC-GRU | 5.25 | 8.56 | 12.45 | 5.99 | 10.37 | 14.18 | 7.32 | 13.47 | 17.11 |
| | DMSTGCN | 2.85 | 5.54 | 7.54 | 3.26 | 6.56 | 9.19 | 3.72 | 7.55 | 10.96 |
| | STSGCN | 2.81 | 5.42 | 7.34 | 3.18 | 6.41 | 8.78 | 3.62 | 7.53 | 10.57 |
| | AGCRN | 3.35 | 7.72 | 8.38 | 4.05 | 9.58 | 10.25 | 4.97 | 11.74 | 12.62 |
| | Z-GCNETs | 3.23 | 7.48 | 7.87 | 3.93 | 9.40 | 9.75 | 4.83 | 11.57 | 12.04 |
| | PGCN | 2.70 | 5.16 | 6.98 | 3.08 | **6.22** | 8.38 | 3.54 | **7.36** | **9.94** |
| Urban-core | HA | 3.05 | 4.68 | 11.77 | 3.47 | 5.16 | 13.69 | 4.03 | 5.83 | 16.26 |
| | FC-LSTM | 2.73 | 4.02 | 11.20 | 2.74 | 4.03 | 11.24 | 2.80 | **4.10** | 11.51 |
| | DCRNN | 2.44 | 3.74 | 9.53 | 2.68 | 4.01 | 10.76 | 2.90 | 4.29 | 11.86 |
| | Graph WaveNet | 2.56 | 3.88 | 9.88 | 2.89 | 4.26 | 11.47 | 3.27 | 4.72 | 13.27 |
| | TGC-GRU | 2.76 | 4.05 | 11.30 | 2.79 | 4.09 | 11.44 | 2.82 | 4.13 | 11.63 |
| | DMSTGCN | 2.42 | **3.70** | 9.42 | 2.62 | **3.94** | 10.44 | 2.81 | 4.17 | 11.37 |
| | STSGCN | 2.44 | 3.73 | 9.53 | 2.63 | 3.96 | 10.50 | 2.79 | 4.17 | 11.29 |
| | AGCRN | 2.44 | 3.74 | 9.57 | **2.60** | **3.94** | 10.43 | **2.77** | 4.16 | 11.24 |
| | Z-GCNETs | 2.44 | 3.74 | 9.57 | 2.63 | 3.96 | 10.57 | 2.83 | 4.20 | 11.52 |
| | PGCN | **2.39** | **3.70** | **9.27** | **2.60** | **3.94** | **10.31** | 2.78 | 4.15 | **11.16** |
| Seattle-Loop | HA | 3.54 | 5.84 | 8.83 | 4.22 | 7.34 | 11.30 | 5.42 | 9.61 | 15.78 |
| | FC-LSTM | 3.72 | 6.02 | 10.95 | 3.92 | 6.46 | 11.70 | 4.25 | 7.12 | 13.11 |
| | DCRNN | **2.83** | **4.76** | **7.38** | **3.28** | 5.84 | **9.38** | 3.86 | 7.07 | 12.01 |
| | Graph WaveNet | 3.10 | 5.11 | 8.35 | 3.68 | 6.37 | 10.83 | 4.50 | 7.94 | 14.55 |
| | TGC-GRU | 3.78 | 6.08 | 10.86 | 4.03 | 6.67 | 11.86 | 4.39 | 7.44 | 13.46 |
| | DMSTGCN | 3.05 | 5.18 | 8.84 | 3.60 | 6.32 | 10.86 | 4.42 | 7.77 | 14.35 |
| | STSGCN | 2.92 | 4.91 | 7.79 | 3.36 | 5.97 | 9.65 | 3.91 | 7.09 | 12.04 |
| | AGCRN | 2.94 | 4.99 | 8.03 | 3.55 | 5.98 | 9.79 | 3.84 | 7.03 | **11.81** |
| | PGCN | 2.85 | 4.80 | 7.56 | **3.28** | **5.80** | 9.46 | **3.82** | **6.94** | 11.86 |
| PEMS04 | HA | 24.92 | 39.00 | 16.29 | 30.28 | 45.87 | 20.26 | 42.38 | 61.67 | 29.96 |
| | FC-LSTM | 21.85 | 33.98 | 14.44 | 25.91 | 39.64 | 17.32 | 34.80 | 51.52 | 24.49 |
| | DCRNN | 19.94 | 31.40 | 13.37 | 22.59 | 35.16 | 15.17 | 27.78 | 42.35 | 19.00 |
| | Graph WaveNet | 18.17 | 28.98 | 13.14 | 19.18 | 30.54 | 13.72 | 21.04 | 33.14 | 15.11 |
| | TGC-GRU | 21.47 | 34.72 | 15.88 | 21.70 | 34.98 | 16.17 | 22.30 | 35.56 | 17.01 |
| | DMSTGCN | 17.62 | 28.49 | 12.22 | **18.41** | **29.78** | **12.61** | **20.00** | **31.99** | 13.81 |
| | STSGCN | 18.56 | 29.96 | 12.58 | 19.46 | 31.53 | 13.10 | 20.98 | 33.76 | 14.42 |
| | AGCRN | 18.65 | 30.56 | 12.42 | 19.35 | 31.75 | 12.83 | 20.82 | 34.13 | **13.68** |
| | Z-GCNETs | 18.64 | 30.20 | 12.36 | 19.36 | 31.48 | 12.79 | 21.00 | 34.09 | 13.72 |
| | PGCN | **17.58** | **28.40** | **12.05** | 18.46 | 29.84 | 12.76 | **20.00** | 32.02 | 13.96 |
| PEMS08 | HA | 19.57 | 29.79 | 11.97 | 24.25 | 36.66 | 14.93 | 34.68 | 50.46 | 21.67 |
| | FC-LSTM | 17.61 | 26.98 | 10.97 | 21.10 | 32.52 | 12.73 | 28.98 | 43.27 | 18.05 |
| | DCRNN | 15.39 | 24.05 | 9.83 | 17.45 | 27.46 | 11.25 | 21.41 | 33.43 | 13.96 |
| | Graph WaveNet | **13.22** | **21.30** | 8.73 | 14.06 | 23.07 | 9.28 | **15.19** | **25.17** | 10.31 |
| | TGC-GRU | 18.90 | 29.35 | 13.00 | 19.47 | 30.27 | 13.41 | 20.57 | 31.90 | 13.99 |
| | DMSTGCN | 13.32 | 21.58 | **8.62** | 14.13 | 23.30 | 9.26 | 15.60 | 25.75 | 10.60 |
| | STSGCN | 14.22 | 22.73 | 9.10 | 15.03 | 24.59 | 9.89 | 16.48 | 27.03 | 11.10 |
| | AGCRN | 14.69 | 23.16 | 9.45 | 15.62 | 24.87 | 9.98 | 17.52 | 27.86 | 11.19 |
| | Z-GCNETs | 15.13 | 23.95 | 9.52 | 16.03 | 25.51 | 10.05 | 17.83 | 28.39 | 11.07 |
| | PGCN | 13.25 | 21.35 | 8.78 | **14.05** | **23.01** | **9.21** | 15.26 | 25.19 | **10.02** |
| PEMS07 | HA | 26.99 | 41.11 | 11.54 | 34.07 | 50.82 | 15.42 | 49.32 | 71.38 | 24.93 |
| | FC-LSTM | 29.49 | 51.46 | 13.12 | 29.68 | 51.71 | 13.20 | 30.16 | 52.02 | 13.48 |
| | DCRNN | 21.03 | 32.83 | 8.92 | 23.94 | 37.05 | 10.28 | 29.49 | 44.59 | 13.07 |
| | Graph WaveNet | 18.70 | **30.54** | **7.90** | 20.12 | 33.04 | 8.64 | 22.48 | 36.57 | 9.67 |
| | TGC-GRU | 29.10 | 48.81 | 13.22 | 29.60 | 49.76 | 13.44 | 30.28 | 50.81 | 13.81 |
| | DMSTGCN | 18.92 | 30.88 | 8.05 | 20.33 | 33.35 | 8.63 | 22.70 | 36.92 | 9.71 |
| | STSGCN | 19.10 | 31.56 | 8.00 | 20.60 | 34.41 | 8.64 | 23.05 | 38.40 | 9.83 |
| | AGCRN | 19.41 | 31.93 | 8.20 | 20.69 | 34.54 | 8.70 | 22.40 | 37.63 | **9.49** |
| | Z-GCNETs | 19.87 | 31.94 | 8.52 | 21.33 | 34.35 | 9.12 | 23.59 | 37.91 | 10.15 |
| | PGCN | **18.68** | 30.55 | 8.08 | **20.02** | **32.97** | **8.49** | **22.19** | **36.27** | 9.57 |

* **bold letters** indicate the best performance for each performance metric in each time step and dataset.

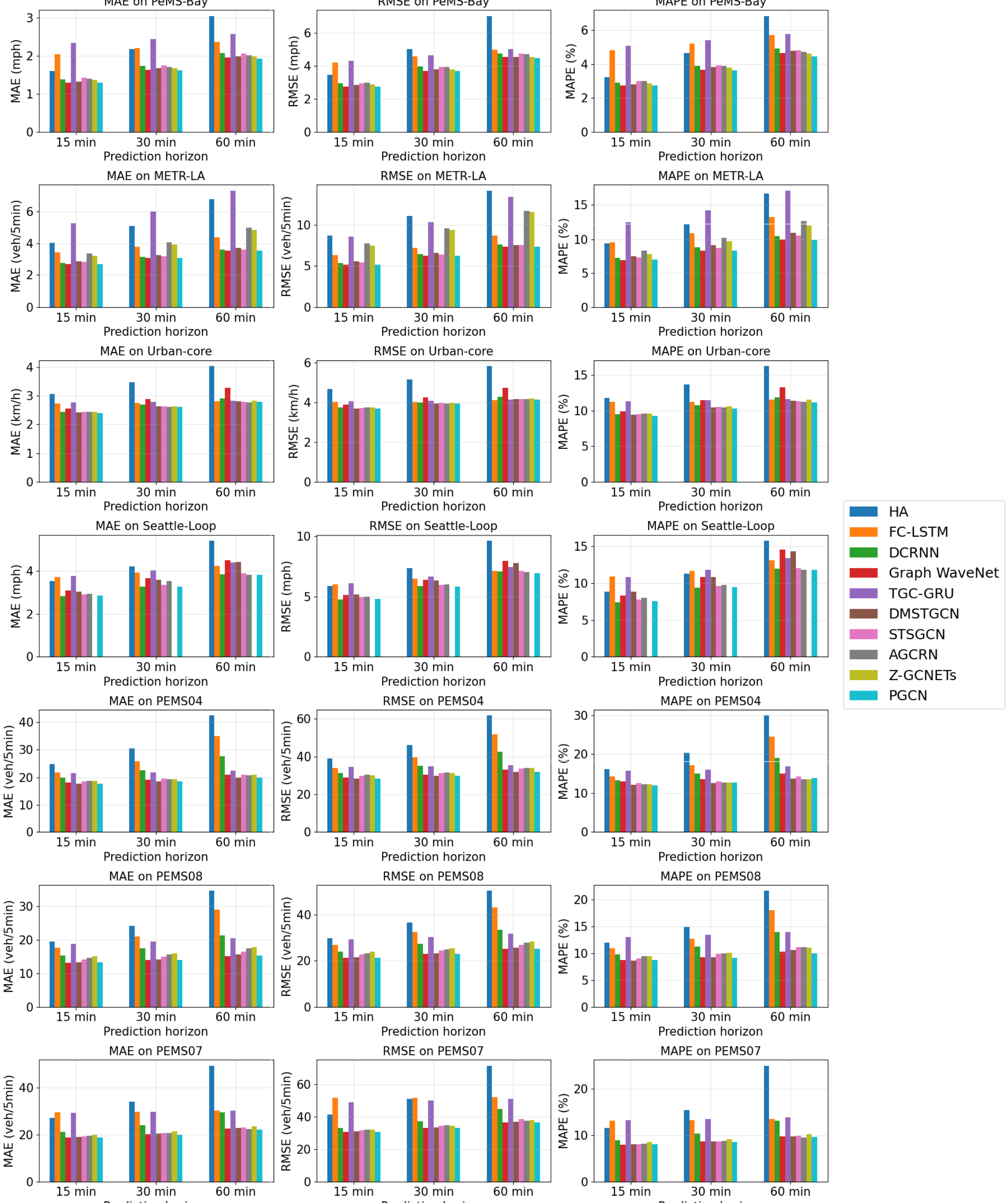

Fig. 5. Performance of PGCN and the baseline models on the seven datasets. PGCN achieves the best or the runner-up performance for all metrics in all prediction horizons and datasets, except for 15-min MAPE on PEMS08 and 60-min RMSE on Urban-core where the model achieves the 3rd best performances, and for 60-min MAPE on PEMS04 and 15-min MAPE on PEMS07 where the model records the 4th best.

results. In Fig. 5, PGCN is always among the best-performing models, if it is not the best. The proposed model achieves the best or the runner-up performance for all metrics in all prediction horizons and datasets, except for 15-min MAPE on PEMS08 and 60-min RMSE on Urban-core for which the model records the 3rd best performance, and for 60-min MAPE on PEMS04 and 15-min MAPE on PEMS07 where the model records the 4th best. Such consistency is observed only for PGCN, where the other models fail to keep competitive performances on at least one dataset. Graph WaveNet shows the highest accuracy on METR-LA, but it performs poorly on Urban-core and Seattle-Loop. On Urban-core, FC-LSTM and TGC-GRU achieve lower RMSE than PGCN for 60-min prediction, but their prediction performances in other datasets are much lower than PGCN's. AGCRN achieves the best performance for 60-min predictions for different metrics on Urban-core, Seattle-Loop, and PEMS04. However, the model shows performances lower by more than 5% compared to PGCN on other tasks. DCRNN and DMSTGCN also achieve the best performances on at least one task, but DCRNN yields lower performances on PeMS-Bay, PEMS04, PEMS08, and PEMS07, and DMSTGCN fails to keep the competitiveness on METR-LA and Seattle-Loop. The performance difference between PGCN and the best-performing model always falls below 1.5%. The only two exceptions are 15-min prediction MAPE on Seattle-Loop where DCRNN outperforms PGCN by 2.44%, and 60-min prediction MAPE on PEMS04 where AGCRN outperforms PGCN by 2.05%. This indicates that constructing graphs that can adapt to data during the testing phase is important in forecasting traffic states to achieve competitive performances with consistency in different application sites. The results demonstrate that the ability of PGCN to adapt to real-time data helps the model obtain robustness against unexpected changes and irregularities in time series compared to the other models with adaptive graphs.

*2) Evaluation of Progressive Graph Convolution:* We conducted an ablation study on the graph convolution module to verify the performance of the proposed model. We tested graph convolution operations with different combinations of adjacency matrices from 3 graph structures, which were transition (T) matrix, self-adaptive (SA) adjacency matrix [13], and progressive (P) adjacency matrix.

Table III shows the MAPE of the different graph convolution modules on 15, 30, and 60-min predictions across all datasets. On PeMS-Bay, Urban-core, PEMS04, and PEMS07, the combination of the transition matrix and progressive adjacency matrix showed the best performances in all prediction horizons except for 15-min on PEMS07. On METR-LA, the combination of the transition matrix and self-adaptive adjacency matrix (Graph WaveNet) outperformed the transition-$p$-graph combination (PGCN) for 15-min and 30-min predictions. On Seattle-Loop, and PEMS08, the combination of using all three types of adjacency matrices produces the best outcome in all prediction horizons except for 30-min prediction on PEMS08 where $p$-graph-self-adaptive combination achieves the state-of-the-art. The results show again that no model can outperform the others in every scenario. However, the combination of the transition matrix and progressive adjacency matrix produces the most consistent results across the seven datasets. Although the $p$-graph alone model does not yield competitive performances on most datasets, it provides complementary information to the transition and self-adaptive adjacency matrices, leading to improved and consistent performance. Also, it is notable that the model achieves competitive performances when two adaptive adjacency matrices are used together. When the graph structure is unavailable, the implementation of the progressive and self-adaptive matrices could offer a sound alternative to traffic forecasting tasks.

TABLE III

RESULT OF ABLATION STUDY OF DIFFERENT ADAPTIVE ADJACENCY MATRICES

| | Graph Convolution | MAPE (%) | | |
|---|---|---|---|---|
| | | 15-min | 30-min | 60-min |
| PeMS-Bay | T | **2.72** | 3.69 | 4.63 |
| | P | 2.90 | 4.11 | 5.72 |
| | SA | 2.81 | 3.85 | 4.85 |
| | P + SA | 2.88 | 4.09 | 5.73 |
| | T + SA | 2.73 | 3.67 | 4.63 |
| | T + P (PGCN) | **2.72** | **3.63** | **4.55** |
| | T+ P + SA | 2.73 | 3.74 | 4.73 |
| METR-LA | T | 6.98 | 8.43 | 10.16 |
| | P | 7.84 | 10.03 | 13.24 |
| | SA | 7.34 | 8.74 | 10.14 |
| | P + SA | 7.43 | 8.90 | 10.33 |
| | T + SA | **6.90** | **8.37** | 10.01 |
| | T + P (PGCN) | 6.98 | 8.38 | **9.94** |
| | T + P + SA | 7.05 | 8.48 | 10.13 |
| Urban-core | T | 9.41 | 10.58 | 11.59 |
| | P | 9.78 | 11.37 | 13.28 |
| | SA | 9.29 | 10.36 | 11.21 |
| | P + SA | 9.33 | 10.39 | 11.25 |
| | T + SA | 9.88 | 11.47 | 13.27 |
| | T + P (PGCN) | **9.27** | **10.31** | **11.16** |
| | T + P + SA | **9.27** | 10.36 | 11.18 |
| Seattle-Loop | T | 7.69 | 9.79 | 12.47 |
| | P | 8.10 | 10.44 | 14.09 |
| | SA | 7.62 | 9.36 | 11.37 |
| | P + SA | 7.80 | 9.69 | 12.05 |
| | T + SA | 8.35 | 10.83 | 14.55 |
| | T + P (PGCN) | **7.56** | 9.46 | 11.86 |
| | T + P + SA | **7.56** | **9.20** | **11.13** |
| PEMS04 | T | 12.92 | 14.10 | 15.88 |
| | P | 12.76 | 13.91 | 16.37 |
| | SA | 12.56 | 13.30 | 14.53 |
| | P + SA | 12.34 | 13.19 | 14.59 |
| | T + SA | 13.14 | 13.72 | 15.11 |
| | T + P (PGCN) | **12.05** | **12.76** | **13.96** |
| | T + P + SA | 12.14 | 12.80 | 14.07 |
| PEMS08 | T | 8.88 | 9.33 | 10.48 |
| | P | 8.69 | 9.44 | 10.71 |
| | SA | 8.67 | 9.13 | 10.25 |
| | P + SA | 8.72 | **9.06** | 10.09 |
| | T + SA | 8.73 | 9.28 | 10.31 |
| | T + P (PGCN) | 8.78 | 9.21 | 10.02 |
| | T + P + SA | **8.66** | 9.14 | **9.81** |
| PEMS07 | T | 8.15 | 8.81 | 10.32 |
| | P | 8.36 | 9.36 | 11.43 |
| | SA | 7.97 | 8.80 | 9.84 |
| | P + SA | 8.01 | 8.68 | 9.79 |
| | T + SA | **7.90** | 8.64 | 9.67 |
| | T + P (PGCN) | 8.08 | **8.49** | **9.57** |
| | T + P + SA | 8.00 | 8.67 | 9.86 |

T: Transition matrix
P: Progressive adjacency matrix
SA: Self-adaptive adjacency matrix [13]

In Fig. 6, the speeds of two sensors in the PeMS-Bay dataset and the 12-step moving average of weights between them are illustrated for a day. Before the morning peak begins for sensor 401936 at 9 am, two nodes share similar traffic trends. The weights between them are not very high at this time because most of the nodes share similar traffic trends (free-flow) before the morning peak begins. Due to the softmax function applied to progressive adjacency matrices as in Eq. (3), the weight becomes smaller as the number of nodes sharing similar traffic trends increases. However, the learned similarity value between the two sensor stations projects high as the speed trend becomes almost identical between 7:30 am

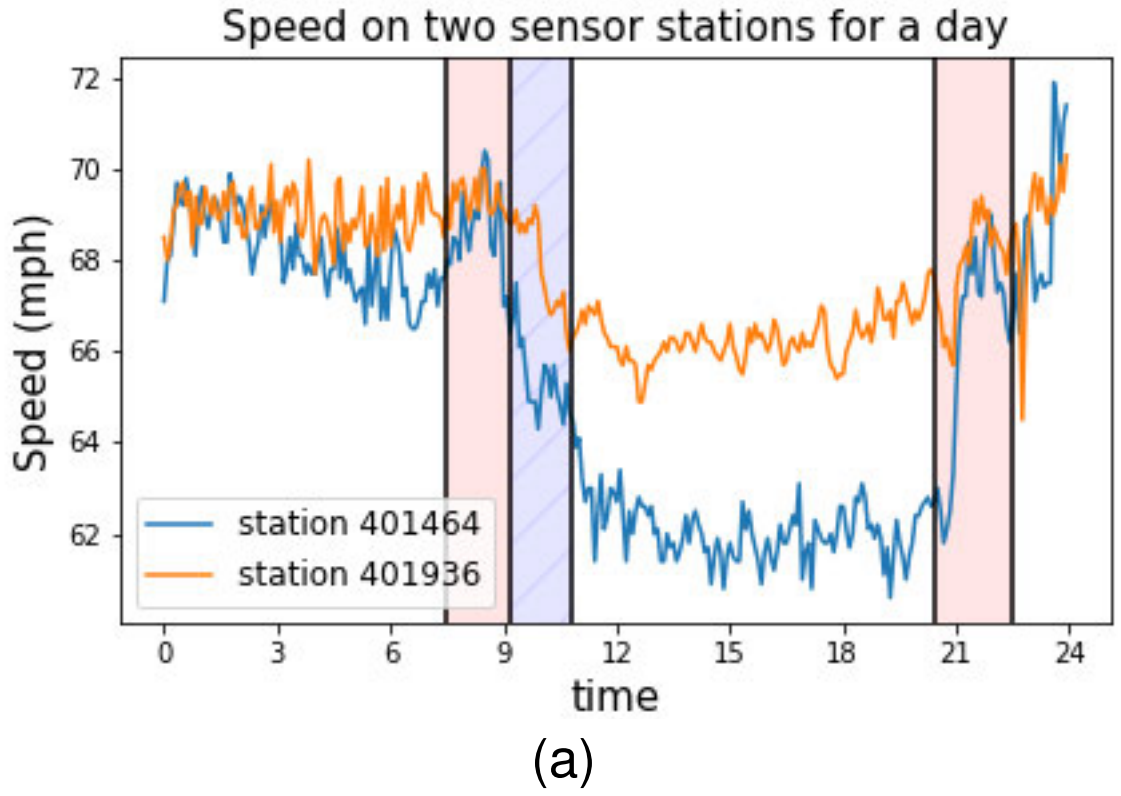

(a)

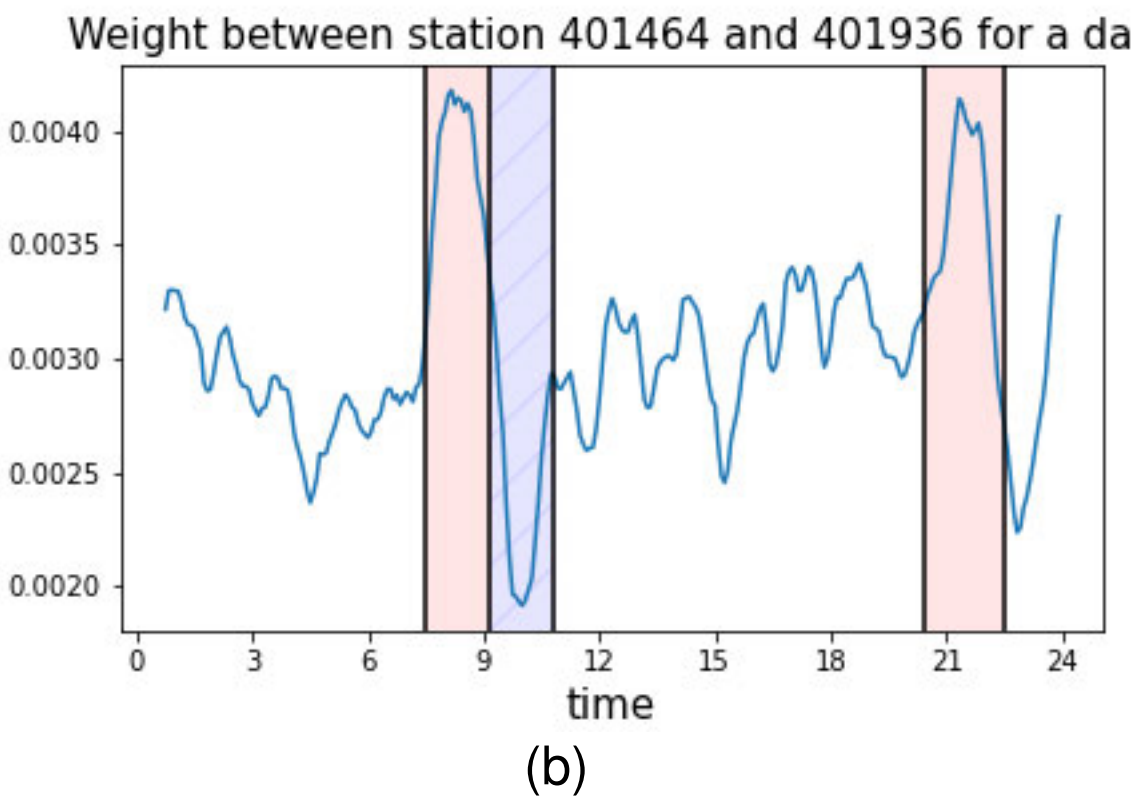

(b)

Fig. 6. (a) Traffic speed and (b) weights for two stations in PeMS-Bay (averaged over 12 time steps). The weight is large when the speed trends are similar (red area), and small when the trends of the signals deviate from each other (blue area with diagonals).

TABLE IV

RESULT OF FEATURE SELECTION RESULTS FOR PEMS04 AND PEMS08

| | Features | MAPE (%) 15-min | 30-min | 60-min |
|---|---|---|---|---|
| PEMS04 | F | **12.05** | **12.76** | **13.96** |
| | F & S | 12.44 | 13.17 | 14.69 |
| | F & O | 12.12 | 12.82 | 14.31 |
| | F, S & O | 12.28 | 12.99 | 14.07 |
| PEMS08 | F | 8.78 | 9.21 | **10.02** |
| | F & S | 8.98 | 9.41 | 10.66 |
| | F & O | **8.50** | **9.02** | 10.03 |
| | F, S & O | 8.76 | 9.21 | 10.23 |

F: Flow, S: Speed, O: Occupancy

and 9 am, and drops significantly with the differences in the rate of speed change after 9 am. In addition, the two nodes share similar traffic trends during the afternoon although the actual speed values are different. Counting on the similarity of the traffic trend, the similarity values are even higher than the period before the morning peak. Finally, when the nodes coincidentally recover the free-flow speed, the similarity value increases by a large margin. This demonstrates the ability of the proposed model to progressively adapt to the data used for the task, and capture the spatial correlations between nodes in transportation networks.

*3) Feature Selection for Multi-Feature Data:* For multi-feature datasets, we conduct an experiment to find the combination of input information in the Progressive Graph Convolution that yields the best outcome. In the scope of our study, the problem is to predict future traffic flow using historical observation of three measurements, flow, occupancy, and speed, of traffic states on different nodes. To include information from multi-feature datasets, we put additional $p$-graph terms to Eq. (5), using progressive adjacency matrices constructed by different features. In mathematical terms,

$$\boldsymbol{Z}_t = \boldsymbol{X}_t \star_{\mathcal{G}} f_{\boldsymbol{W}} = \sum_{k=0}^{K-1} \Big( \boldsymbol{P}^k \boldsymbol{X}_t \boldsymbol{W}_{k,1} + \boldsymbol{P}^{\top^k} \boldsymbol{X}_t \boldsymbol{W}_{k,2} + \boldsymbol{A}^t_{P_{tgt}} \boldsymbol{X}_t \boldsymbol{W}_{k,3} + \boldsymbol{A}^t_{P_{aux}} \boldsymbol{X}_t \boldsymbol{W}_{k,4} \Big), \tag{11}$$

for utilizing two features, where $A^t_{P_{tgt}}$ is the progressive adjacency matrix constructed using the historical target measure (flow) at time $t$, and $A^t_{P_{aux}}$, using one of historical auxiliary measure (speed *or* occupancy), and

$$\boldsymbol{Z}_t = \boldsymbol{X}_t \star_{\mathcal{G}} f_{\boldsymbol{W}} = \sum_{k=0}^{K-1} \Big( \boldsymbol{P}^k \boldsymbol{X}_t \boldsymbol{W}_{k,1} + \boldsymbol{P}^{\top^k} \boldsymbol{X}_t \boldsymbol{W}_{k,2} + \boldsymbol{A}^t_{P_{tgt}} \boldsymbol{X}_t \boldsymbol{W}_{k,3} + \boldsymbol{A}^t_{P_{aux_1}} \boldsymbol{X}_t \boldsymbol{W}_{k,4} + \boldsymbol{A}^t_{P_{aux_2}} \boldsymbol{X}_t \boldsymbol{W}_{k,5} \Big), \tag{12}$$

for utilizing three features, where $A^t_{P_{aux_1}}$, and $A^t_{P_{aux_2}}$ are the progressive adjacency matrices constructed using both of historical auxiliary measures (speed *and* occupancy). Table IV shows the results of the feature selection experiment. Since the target measure is traffic flow, we conduct experiment using every combination that includes traffic flow.

The results of our experiments show that for both multi-feature datasets, the single-feature model performs the best in 60-min predictions on both PEMS04 and PEMS08. On PEMS04, the single-feature model yields the most accurate predictions for all horizons. On the other hand, on PEMS08, a two-feature model incorporating flow and occupancy outperforms the single-feature model for 15 and 30-min predictions. Our findings indicate that the single-feature model offers the most robust forecasting with low computational requirements. Additionally, it has the added benefit of eliminating the time-intensive feature selection process, making it efficient for applications on other multi-feature datasets.

*4) Computation Efficiency:* While progressively adapting to the online traffic data, PGCN still captures computation efficiency compared to other spatial-temporal graph neural networks. Among constructions for different adaptive graphs, the progressive adjacency matrix requires the least number of parameters. Self-adaptive adjacency matrix [13] is constructed by multiplying source and target node embedding matrices $\boldsymbol{E}_1$ and $\boldsymbol{E}_2 \in \mathbb{R}^{N \times d}$. DMSTGCN requires large memory space to construct the dynamic graph compared to the other models with adaptive adjacency matrices. The required parameters for DMSTGCN are embedding of time slots $\boldsymbol{E}^t \in \mathbb{R}^{N \times d}$, embeddings of source and target nodes $\boldsymbol{E}_1$ and $\boldsymbol{E}_2 \in \mathbb{R}^{N \times d}$, and a core tensor $\boldsymbol{E}^k \in \mathbb{R}^{d \times d \times d}$, where $N_t$ is the number of time slots in a day (288 for the datasets used in this study). The progressive adjacency matrix only requires a parameter $\boldsymbol{W}_{adj} \in \mathbb{R}^{T \times T}$ for the similarity calculation, where $T$ is the length of the historical traffic states. Yet, PGCN generalizes better than Graph WaveNet and DMSTGCN to more study sites. Table V describes the computation time and the number of total parameters for PGCN, DCRNN, Graph WaveNet, DMSTGCN, STSGCN, AGCRN, and Z-GCNETs on the PeMS-Bay dataset. STSGCN requires large computation capability due to the adjacency matrix for the localized spatial-temporal graph $A' \in \mathbb{R}^{3N\times 3N}$. For Z-GCNETs, extra computation time to generate zigzag persistence images is required, which is about 10 sec for an observation.

TABLE V
THE NUMBER OF PARAMETERS AND COMPUTATION COST ON THE PEMS-BAY DATASET

| Model | # Params | Computation Time | |
|---|---|---|---|
| | | Training (s/epoch) | Inference |
| PGCN | 305,404 | 104.2 | 4.0 |
| DCRNN | 223,744 | 177.1 | 13.4 |
| Graph WaveNet | 311,760 | 103.6 | 4.6 |
| DMSTGCN | 351,692 | 59.1 | 2.0 |
| STSGCN | 2,045,997 | 382.3 | 15.3 |
| AGCRN | 748,990 | 99.3 | 4.3 |
| Z-GCNETs | 455,214 | 131.7 | 6.7 |

## VI. CONCLUSION AND FUTURE STUDY

In this paper, we proposed a spatial-temporal traffic forecasting model, Progressive Graph Convolutional Networks (PGCN). The proposed model captures the time-varying spatial correlations by progressively adapting to data used for forecasting tasks. Instead of finalizing adaptive graphs after the training phase, we used parameterized cosine similarity of traffic measures to make the graph adjust dynamically to the online traffic data even during the testing phase. The experimental results show that the model with a progressive graph can consistently achieve state-of-the-art performance in all seven of the datasets used in this study. The model showed the ability to generalize in both single-feature and multi-feature datasets, proving the necessity of reflecting online input data to acquire robustness.

In current model, the weight parameters for the temporal feature extraction module remain fixed for every input. Since the temporal correlations between time steps may also change, we aim to introduce a dynamic weight assignment mechanism for the temporal dimension to enhance traffic forecasting accuracy. While self-attention [43] assigns weights based on online input data, it overlooks the crucial context that close time steps have a more substantial impact on predictions for the near future [44]. Meanwhile, causal convolutions and RNNs assign weights based on sequential information. We will develop a temporal feature extraction module that captures the advantages of both self-attention and causal convolution to assign weights based on online traffic data while preserving sequential information.

The second direction is improving the graph convolution module within our PGCN to effectively incorporate multiple graph structures. Currently, we implement the diffusion convolution approach which sums up the representations extracted from different graphs. This may dilute information extracted from various graph structures, potentially compromising the model performance. To address this limitation, we will develop a novel graph convolution module that efficiently aggregates representations without information dilution.

Finally, we will incorporate external features of transportation systems such as weather, road characteristics, and POI (Point-of-Interests) into the model. In this study, we have demonstrated that the model has achieved robust performance even without the information about the structural information of the transportation network. In the later work, we will explore how the external features influence the performance of traffic forecasting across diverse study sites.

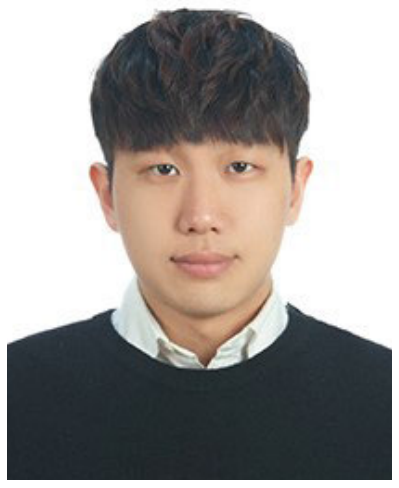

**Yuyol Shin** (Member, IEEE) received the B.S. and Ph.D. degrees in civil and environmental engineering from the Korea Advanced Institute of Science and Technology (KAIST), Daejeon, South Korea, in 2016 and 2022, respectively. He is currently a Post-Doctoral Researcher of civil and environmental engineering with KAIST and a Visiting Scholar of civil engineering with UC Berkeley. His research interests include spatio-temporal data mining, artificial intelligence, and transportation network analysis.

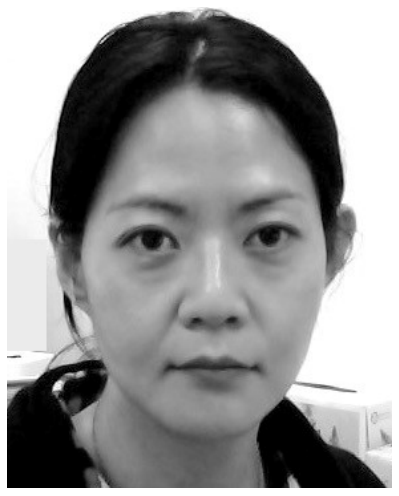

**Yoonjin Yoon** (Member, IEEE) received the B.S. degree in mathematics from Seoul National University, Seoul, South Korea, in 1996, and the M.S. degree in computer science and the M.S. degree in management science and engineering from Stanford University, Stanford, CA, USA, in 2000 and 2002, respectively, and the Ph.D. degree in civil and environmental engineering from the University of California at Berkeley, Berkeley, CA, USA, in 2010. Since 2011, she has been an Assistant Professor with the Department of Civil and Environmental Engineering, Korea Advanced Institute of Science and Technology (KAIST), Daejeon, South Korea. Before arriving at KAIST, she was a Graduate Student Researcher with the National Center of Excellence in Air Transportation Operations Research (NeXTOR), University of California at Berkeley, from 2005 to 2010. Previously, she was a Research Assistant with the Artificial Intelligence Center, SRI International, Menlo Park, CA, USA, in 1999, and the Center of Reliability Computing, Stanford University, Stanford, CA, USA, from 2000 to 2002. Her research interests include the traffic management of both manned and unmanned vehicles using stochastic optimization, data-driven driving behavior analysis, and autonomous vehicle traffic flow management using large-scale driving data.